\documentclass[12pt,leqno,oneside]{amsart}
\nonstopmode
\usepackage{egame}
\usepackage{natbib}

\begin{document}

\vfuzz2pt \hfuzz2pt

\newtheorem{theorem}{Theorem}
\newtheorem{prop}{Proposition}
\newtheorem{lemma}{Lemma}
\newtheorem{claim}{Claim}

\newtheorem{axiom}{Axiom}
\newtheorem{definition}{Definition}
\newtheorem{property}{Property}
\newtheorem{observation}{Observation}
\newtheorem*{stability}{Stability Theorem}
\newtheorem*{myopic}{The myopic strategy rule}
\newtheorem*{memoryless}{The memoryless revision rule}
\newtheorem*{exploratory}{The $\delta$-exploratory myopic strategy rule}
\newtheorem*{averaging}{The averaging revision rule}

\theoremstyle{definition}
\newtheorem{example}{Example}

\def\st{such that\ }
\def\proof#1{\medskip\noindent{\bf {Proof{#1}.\  }}}
\def\qed{\vrule height6pt width4pt depth0pt}

\def\P{\Phi}
\def\b{\beta}
\def\p{\phi}
\def\O{\Omega}
\def\o{\omega}
\def\S{\Sigma}
\def\s{\sigma}
\def\m{\mu}
\def\x{\xi}
\def\d{{\delta}}
\def\n{{\nu}}
\def\l{\lambda}
\def\a{\alpha}
\def\D{\Delta}
\def\e{\varepsilon}
\def\f{\varphi}

\begin{abstract}
A valuation for a player in a game in extensive form is an
assignment of numeric values to the players moves. The valuation
reflects the desirability moves. We assume a myopic player, who
chooses a move with the highest valuation. Valuations can also be
revised, and hopefully improved, after each play of the game.
Here,  a very simple valuation revision is considered, in which
the moves made in a play are assigned the payoff obtained in the
play. We show that  by adopting such a learning process a player
who has a winning strategy in a win-lose game can almost surely
guarantee a win in a repeated game. When a player has more than
two payoffs, a more elaborate learning procedure is required. We
consider one that associates with each move the average payoff in
the rounds in which this move was made. When all players adopt
this learning procedure, with some perturbations, then, with
probability 1, strategies that are close to subgame perfect
equilibrium are played after some time. A single player who adopts
this procedure can guarantee only her individually rational
payoff.
\end{abstract}

\title{Learning to play games in extensive form by valuation}

\author{Philippe Jehiel and Dov Samet}
\date{November, 2000}
\maketitle

\section{Introduction}

Models of learning in games fall roughly into two categories. In
the first, the learning player forms beliefs about the future
behavior of other players and nature, and directs her behavior
according to these beliefs. We refer to these as
fictitious-player-like models. In the second, the player is
attuned only to her own performance in the game, and uses it to
improve future performance. These are called models of
reinforcement learning.

Reinforcement learning has been used extensively in artificial
intelligence (AI). Samuel wrote a checkers-playing learning
program as far back as 1955, which marks the beginning of
reinforcement learning (see \cite{samuel}). Since then many other
sophisticated algorithms, heuristics, and computer programs, have
been developed, which are based on reinforcement learning.
(\cite{suttonbarto}). Such programs try neither to learn the
behavior of a specific opponent, nor to find the distribution of
opponents' behavior in the population. Instead, they learn how to
improve their play from the achievements of past behavior.

Until recently, game theorists studied mostly
fictitious-player-like models. Reinforcement learning has only
attracted  the attention of game theorists  in the last decade in
theoretical works like \cite{gilbsch}, \cite{camererho},
\cite{sarinvahid}, and in experimental works like \cite{erevroth}.
In all these studies the basic model is given in a strategic form,
and the learning player identifies those of her strategies that
perform better. This approach seems inadequate where learning of
games in extensive form is concerned. Except for the simplest
games in extensive form, the size of the strategy space is so
large that learning, by human beings or even machines, cannot
involve the set of all strategies. This is certainly true for the
game of chess, where the number of strategies exceeds the number
of particles in the universe. But even a simple game like
tic-tac-toe is not perceived by human players in the full extent
of its strategic form.

The process of learning games in extensive form can involve only a
relatively small number of simple strategies. But when the
strategic form is the basic model, no subset of strategies can be
singled out. Thus, for games in extensive form the structure of
the game tree should be taken into consideration. Instead of
\emph{strategies} being reinforced, as for games in strategic
form, it is the \emph{moves} of the game that should be reinforced
for games in extensive form.

This, indeed, is the approach of heuristics for playing games
which were developed by AI theorists.\footnote{Perhaps the
concentration of the AI literature on moves rather than strategies
is the reason why there seems to be almost no overlap between two
major books on learning, each in its field: {\em The Theory of
Learning in Games\/}, \cite{fudelevine} and {\em Reinforcement
Learning: An Introduction\/}, \cite{suttonbarto}. } One of the
most common building block of such heuristics is the
\emph{valuation}, which is a real valued function on the possible
moves of the learning player. The valuation of a move reflects,
very roughly, the desirability of the move. Given a valuation, a
learning process can be defined by specifying two rules:

\begin{itemize}
  \item A \emph{strategy rule}, which specifies  how the game is played
   for any given  valuation of the player;
  \item A \emph{revision rule}, which specifies how the valuation
  is revised after playing the game.
\end{itemize}

Our purpose here is to study learning-by-valuation processes,
based on simple strategy and revision rules. In particular, we
want to demonstrate the convergence properties of these processes
in repeated games, where the stage game is given in an extensive
form with perfect information and any number of players.
Converging results of the type we prove here are very common in
the literature of game theory. But as noted before, convergence of
reinforcement is limited in this literature to strategies rather
than moves.\footnote{There is no obvious way to define an
assessment for a strategy from a system of node valuations.
Therefore, a simple translation of our learning model in terms of
strategies is not straightforward. One fundamental difficulty is
that the node valuation treatment does not impose that a strategy
be assessed in the same way throughout the play of the game. Also,
two strategies involving the same first move should be assessed in
the same way initially (a condition which does not make much sense
in the reinforcement learning based on the strategic form.} To the
best of our knowledge, the AI literature while describing dynamic
processes closely related to the ones we study here do not prove
convergence results of this type.

First, we study stage games in which the learning player has only
two payoffs, 1 (win) and 0 (lose). Two-person win-lose games are a
special case. But here, there is  no restriction on the number of
the other players or their payoffs.

For these games we adopt the simple \emph{ myopic strategy rule}.
By this rule, the player chooses in each of her decision node a
move which has the highest valuation among the moves available to
her at this node. In case there are several moves with the highest
valuation, she chooses one of them at random.

As a revision rule we adopt the simple \emph{memoryless revision}:
after each round the player revises  only the valuation of the
moves  made in the round. The valuation of such a move becomes the
payoff (0 or 1) in that round.

Equipped with these rules, and an initial valuation, the player
can play a repeated game. In  each round she plays according to
the myopic strategy, using the current valuation, and at the end
of the round she revises her valuation according to the memoryless
revision.

This learning process, together with the strategies of the other
players in the repeated game, induce a probability distribution
over the infinite histories of the repeated game. We show the
following, with respect to this probability.

\begin{quote}
Suppose that the learning player can guarantee a win in the stage
game. If  she plays according to the  myopic strategy and the
memoryless revision rules, then starting with any nonnegative
valuation, there exists, with probability 1, a time after which
the player always wins.
\end{quote}

When the learning player has more than two payoffs, the previous
learning process is of no help. In this case we study the
\emph{exploratory myopic strategy rule}, by which the player opts
for the maximally valued move, but  chooses also, with small
probability, moves that do not maximize the valuation.

The introduction of such perturbations makes it necessary to
strengthen the revision rule. We consider the \emph{averaging
revision}. Like the memoryless revision, the player revises only
the valuation of moves made in the last round. The valuation of
such a move is the average of the payoffs in all previous rounds
in which this move was made.

\begin{quote}
If the learning player obeys the exploratory myopic strategy and
the averaging revision rules,  then starting with any valuation,
there exists, with probability 1, a time after which the player's
payoff is close to her individually rational payoff (the maxmin
payoff) in the stage game.
\end{quote}

The two previous results indicate that reinforcement learning
achieves learning of playing the stage game itself, rather than
playing against certain opponents. The learning processes
described guarantee the player her individually rational payoff
(which is the win in the first result). This is exactly the payoff
that she can guarantee even when the other players are
disregarded.

Our next result concerns the case where all the players learn the
stage game. By the previous result we know that each can guarantee
his individually rational payoff. But, it turns out that the
synergy of the learning processes yields the players more than
just learning the stage game. Indeed, they learn in this case each
other's behavior and act rationally on this information.

\begin{quote}
Suppose the stage game has a unique perfect equilibrium. If all
the players employ the exploratory myopic strategy and the
averaging revision rules, then starting with any valuation, with
probability 1, there is a time after which their strategy in the
stage game is close to the perfect equilibrium.
\end{quote}

Although valuation is defined for all moves, the learning player
needs no information concerning the game when she start playing
it. Indeed, the initial valuation can be constant. To play the
stage game with this valuation, the player needs to know which
moves are possible to her, only  when it is her turn to play, and
then choose one of them at random. During the repeated game, the
player should be able to record the moves she made and their
valuations. Still, the learning procedure does not require that
the player knows how many players there are, let alone the moves
they can make and their payoffs.

The learning processes discussed here treat separately the
valuation for every node. For games with large number of nodes (or
states of the board), that may be unrealistic because the chance
of meeting a given node several times is too small. In chess, for
example, almost any state of the board, except for the few first
ones, has been seen in recorded history only once. In order to
make these processes more practical, similar moves (or states of
the board) should be grouped together, such that the number of
similarity classes is manageable. When the valuation of a move is
revised, so are all the moves similar to it. We will deal with
such learning processes, as well as with games with incomplete
information, in a later paper.

\section{Preliminaries}
\subsection{Games and super games}
Consider a finite game $G$  with complete information and a finite
set of players $I$. The game is described by a tree $(Z,N,r,A)$,
where $Z$ and $N$ are the sets of terminal and non-terminal nodes,
correspondingly, the root of the tree is $r$, and the set of arcs
is $A$. Elements of $A$ are ordered pairs $(n,m)$, where $m$ is
the immediate successor of $n$.

The set $N_i$, for $i\in I$, is the set of nodes in which it is
$i$'s turn to play. The sets $N_i$ form a partition of $N$.  The
\emph{moves} of player $i$ at node $n\in N_i$ are the nodes in
$M_i(n)=\{m\mid (n,m)\in A\}$. Denote $M_i=\cup_{n\in N_i}
M_i(n)$. For each $i$ the function $f_i\colon Z\to R$ is $i$'s
payoff function. The depth of the game is the length of the
longest path in the tree. A game with depth 0 is one in which
$\{r\}=Z$ and $N=\emptyset$.

A behavioral strategy, (strategy for short) for player $i$ is a
function $\s_i$ defined on $N_i$, such that for each $n\in N_i$,
$\s_i(n)$ is a probability distribution on $M_i(n)$.

The super game $\Gamma$ is the  infinitely repeated game, with
stage game $G$. An infinite history  in $\Gamma$ is an element of
$Z^\omega$. A finite history of $t$ rounds, for $t\ge 0$, is an
element of $Z^t$. A {\em super strategy} for player $i$ in
$\Gamma$ is a function $\Sigma_i$ on finite histories, such that
for $h\in Z^t$, $\Sigma_i(h)$ is a strategy of $i$ in $G$, played
in round $t+1$. The super strategy $\Sigma=(\Sigma_i)_{i\in I}$
induces a probability distribution on histories in the usual way.

\subsection {\bf Valuations.} We
fix  one player $i$ (the learning player) and omit subscripts of
this player when the context allows it. We first introduce the
basic notions of playing by  valuation.  A {\em valuation} for
player $i$ is a function $v\colon M_i\to R$.

Playing the repeated game $\Gamma$ by valuation requires two rules
that describe how  the stage game $G$ is played for a given
valuation, and how a valuation is revised after playing $G$.

\begin{itemize}
  \item A \emph{strategy rule} is a function $v\rightarrow \s^v$.
  When player $i$'s valuation is $v$, $i$'s strategy in $G$ is $\s^v$.
  \item A \emph{ revision rule} is a function $(v,h)\rightarrow
  v^h$, such that for the empty history $\Lambda$, $v^\Lambda=v$.
 When player $i$'s initial valuation is $v$, then after a history of plays
 $h$, $i$'s valuation is $v^h$.

\end{itemize}

\begin{definition}
The \emph{valuation super strategy} for player $i$,  induced by a
strategy rule $v\rightarrow \s^v$, a revision rule
$(v,h)\rightarrow v^h$, and an initial valuation $v$, is the super
strategy $\Sigma^{v}_i$, which is defined by
$\Sigma^{v}_i(h)=\sigma^{v^h}$ for each finite history $h$.
\end{definition}

\section{Main results}
\subsection{Win-lose games}

We consider first the case where player $i$ has two possible
payoffs in $G$, which are, without loss of generality, 1 (win) and
0 (lose). A two-person win-lose game is a special case, but here
we place no restrictions on the number of players or their
payoffs.

We assume that learning by valuation is induced by a strategy rule
and a revision rule of a simple form.

\begin{myopic}
This rule associates with each valuation $v$ the strategy
$\s^{v}$, where for each node $n\in N_i$, $\s^{v}(n)$ is the the
uniform distribution over the maximizers of $v$ on $M_i(n)$. That
is, in each node of player $i$, the player selects at random one
of the moves with the highest valuation.
\end{myopic}

\begin{memoryless}
 For a history
$h=(z)$ of length 1, the valuation $v$ is revised to $v^z$ which
is defined for each  node $m\in M_i(n) $ by
 $$ v^z(m)=\begin{cases}
    f_i(z) & \text{$m$ is on the path leading from $r$ to $z$}, \\
   v(m) & \text{otherwise}.
  \end{cases}$$
For a history $h=(z_1,\dots,z_t)$,  the current valuation is
revised in each round  according to the terminal node observed in
this round. Thus, $v^h=\big(v_i^{(z_1,\dots,z_{t-1})}\big)^{z_t}$.
\end{memoryless}

The temporal horizons, future and past, required for these two
rules are very narrow. Playing the game $G$, the player takes into
consideration just her next move. The revision of the valuation
after playing $G$ depends only on the current valuation, and the
result of this play, and not on the history of past valuations and
plays. In addition, the revision is confined only to those moves
that were made in the last round.

\begin{theorem}\label{main}
Let $G$ be a game in which player $i$ either wins  or loses.
Assume that player $i$ has a strategy in $G$ that guarantees him a
win. Then for any initial nonnegative valuation $v$ of $i$, and
super strategies $\Sigma$ in $\Gamma$, if $\Sigma_i$ is the
valuation super strategy induced by the myopic strategy and the
memoryless revision rules, then with probability 1, there is a
time after which $i$ is winning forever.
\end{theorem}

The following  example demonstrates learning by valuation.

\begin{example}
Consider  the game in Figure \ref{f:goodcase}, where  the payoffs
are player 1's.

\begin{figure}[htb]
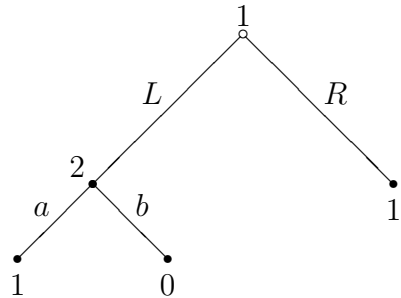

\hspace*{\fill}
\begin{egame}(600,380)
\putbranch(300,340)(1,1){200} \iib{1}{$L$}{$R$}[][$1$]
\putbranch(100,140)(1,1){100} \iib{2}[l]{$a$}{$b$}[$1$][$0$]
\end{egame}
\hspace*{\fill} \caption[]{Two payoffs} \label{f:goodcase}
\end{figure}
\end{example}

Suppose that  1's initial valuation of each of the moves $L$ and
$R$ is 0. The valuations that will follow can be one of  $(0,0)$,
$(1,0)$, and $(0,1)$, where the first number in each pair is the
valuation of $L$ and the second of $R$. (The valuation $(1,1)$
cannot be reached from any of these valuations).

We can think of these possible valuations as states in a
stochastic process. The state $(0,1)$ is absorbing. Once it is
reached, player 1 is choosing $R$ and being paid  1 forever. When
the valuation is $(1,0)$, player 1 goes $L$. She will keep going
$L$, and  winning 1, as long as player 2 is choosing $a$. Once
player 2 chooses $b$,  the valuation goes back to $(0,0)$. Thus,
the only way player 1 can fail to be paid 1 from a certain time on
is when $(0,0)$ recurs infinitely many times. But the probability
of this is 0, as the probability of reaching the absorbing state
$(0,1)$ from state $(0,0)$ is 1/2.

Note that the theorem does not state that with probability 1 there
is a time after which player 1's strategy is the one that
guarantees him payoff 1. Indeed, in this example, if player 2's
strategy is always $a$, then there is a probability 1/2 that
player 1 will play $L$ for ever, which is not the strategy that
guarantees player 1 the payoff 1.

\subsection{The case of payoff function with more than two values}

We now turn to the case in which payoff functions  take more than
two values. The next example shows that in this case the myopic
strategy and the memoryless revision rules may lead the player
astray.

\begin{example}\label{badcase}
Player 1 is the only player in the game in Figure \ref{f:badcase}.

\begin{figure}[htb]
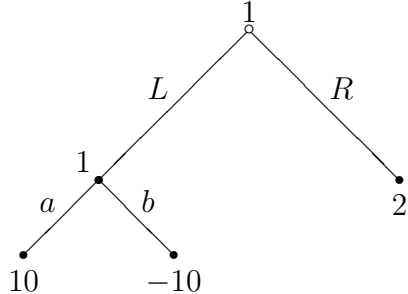

\hspace*{\fill}
\begin{egame}(600,380)
\putbranch(300,340)(1,1){200} \iib{1}{$L$}{$R$}[][$2$]
\putbranch(100,140)(1,1){100} \iib{1}[l]{$a$}{$b$}[$10$][$-10$]
\end{egame}
\hspace*{\fill} \caption[]{More than two payoffs}
\label{f:badcase}
\end{figure}

In this game player 1 can guarantee a payoff of 10, and therefore
we expect a learning process to lead player 1  to this payoff.
But, no reasonable restriction on the initial valuation can
guarantee that the learning process induced by the myopic strategy
and the memoryless revision results in the payoff 10 in the long
run. For example, for any constant initial valuation, there is a
positive probability that the valuation $(-10,2)$ for $(L,R)$ is
obtained, which is absorbing.

We cannot state for general payoff functions any theorem analogous
to Theorem \ref{main} or even a weaker version of this theorem.
But something meaningful can be stated when \emph{all} players
play the repeated game according to the  myopic strategy and the
memoryless revision rules.

We say that game $G$ is \emph{generic} if for every player $i$ and
for every pair of distinct terminal nodes $z$ and  $z^{\prime }$,
we have $f_{i}(z)\neq f_{i}(z^{\prime })$.

\begin{theorem}\label{generic}
Let $G$ be a generic game. Assume that each player $i$ plays
$\Gamma $ according to the myopic strategy rule and uses the
memoryless revision rule. Then for any initial valuation profile,
with probability 1, there is a  time after which the same terminal
node  is reached in each round.
\end{theorem}

The limit plays guaranteed by this theorem  depend on the initial
valuations and have no special structure in  general. Moreover, it
is obvious that for any  terminal node there are initial
valuations that guarantee that this terminal node is reached in
all rounds.

We return, now, to the case where only one player learns by
reinforcement. In order to prevent a player from being paid an
inferior payoff forever, like in Example \ref{badcase}, we change
the strategy rule. We allow for exploratory moves that remind her
of all possible payoffs in the game, so that she is not stuck in a
bad valuation. Assume, then, that having a certain valuation, the
player opts for the highest valued nodes, but still allows for
other nodes with a small probability $\d$. Such a rule guarantees
that player in Example \ref{badcase} will never be stuck in the
valuation $(-10,2)$. We introduce formally this new rule.

\begin{exploratory}
 This rule
associates with each valuation $v$ the strategy $\s^{v}_\d$, where
for each node $n\in N_i$, $\s^{v}_\d(n)=(1-\delta)\sigma^{v}(n)
+\delta \mu(n)$. Here, $\sigma^{v}$ is the strategy associated
with $v$ by the myopic strategy rule, and $\mu$ is the strategy
that uniformly selects one of the moves at $n$.
\end{exploratory}

Unfortunately, adding exploratory moves does not help the player
to achieve 10 in the long run, as we show now. Assume that the
initial valuation of $a$ and $b$ is 10 and $-10$ correspondingly,
and the valuation of the fist two moves is also favorable:
$(10,2)$. We assume now that in each of the two nodes player 1
chooses the higher valued node with probability $1-\d$ and the
other with probability $\d$. The valuation of $a$ and $b$ cannot
change over time. The valuation of $(L,R)$ form  an ergodic Markov
chain with the two states $\{(10,2), (-10,2)\}$. Thus, for
example, the probability of transition from $(10,2)$ to itself
occurs when the player chooses either $L$ and $a$, with
probability $(1-\d)^2$, or $R$ with probability $\d$, which sum to
$1-\d+\d^2$.

The following is the transition matrix of this Markov chain.
$$\bordermatrix{&(10,2)&(-10,2)\cr (10,2)&1-\d+\d^2& \d-\d^2\cr
(-10,2)&\d-\d^2& 1-\d+\d^2\cr}$$ The two states $(10,2)$ and
$(-10,2)$ are symmetric and therefore the stationary probability
of each is 1/2. Thus, the player is paid 10 and 2, half of the
time each.
\end{example}

Note that the exploratory moves are required because the payoff
function has more than two values. However, the failure to achieve
the payoff 10 after introducing the the $\delta$-exploratory
myopic strategy rule is the result of this rule, and has nothing
to do with the number of values of the payoff function. That is,
even in a win-lose game, a player who has a winning strategy may
fail to guarantee a win in the long run by playing according to
the rules of $\delta$-exploratory myopic strategy and memoryless
revision.

Thus, the introduction of the $\delta$-exploratory myopic strategy
rule forces us also  to strengthen  the revision rule as follows.

\begin{averaging}
 For a node
$m\in M_i$, and a  history $h=(z_1,\dots,z_t)$, if the node $m$
was never reached in $h$, then $v^h(m)=v(m)$. Else, let
$t_1,\dots,t_k$ be the times at which $m$ was reached in $h$, then
$$v^h(m)=\frac{1}{k}\sum_{l=1}^k f(z_{t_l}).$$
\end{averaging}

We state, now, that by using little exploration, and averaging
revision, player $i$ can guarantee to be close to his individually
rational (maxmin) payoff in $G$.

\begin{theorem}\label{maxmin}
Let $\S$ be a super strategy such that $\S_i$ is the valuation
super strategy induced by the $\delta$-exploratory myopic strategy
and the averaging revision rules. Denote by $P_\delta$  the
distribution over histories in $\Gamma$ induced by $\Sigma$.

Let $\rho$ be $i$'s individually rational payoff in $G$. Then for
every $\e>0$ there exists $\delta_0>0$ such that for every
$0<\delta<\delta_0$, for $P_\delta$-almost all infinite histories
$h=(z_1,z_2,\dots)$,
 $$\varliminf_{t\rightarrow\infty}\,\frac{1}{t}\sum_{l=1}^t f(z_l)>\rho-\e.$$
\end{theorem}

We consider now the case where all players learn to play $G$,
using the $\d$-exploratory myopic strategy and the averaging
revision rules. We show that in such a case,  in the long run, the
players' strategy in the stage game is close to a perfect
equilibrium. We assume for simplicity that the game $G$ has a
unique perfect equilibrium (which is true generically).

\begin{theorem}\label{perfect}
Assume that $G$ has a unique perfect equilibrium
$\beta=(\beta_i)_{i\in I}$. Let $\S^\d$ be the super strategy such
that for each $i$, $\S_i^\d$ is the valuation super strategy
induced by the $\d$-exploratory myopic strategy, and the averaging
revision rules.

Let $P_\d$ be the distribution over histories induced by $\S^\d$.
Then there exists $\d_0$, such that for all $0<\d<\d_0$, for
$P_\d$-almost all infinite histories $h=(z_1,\dots,z_t,\dots)$,
there exists $T$, such that for all $t>T$,
$\s_i^{v^{(z_1,\dots,z_t)}}(m)=(1-\d)\b_i(m) +\d\m(m)$, for each
player $i$ and node $m\in M_i$.
\end{theorem}

\section{proofs}
\subsection{Stochastic repeated games} We prove all the theorems by
induction on the depth of the game tree. For this we need to be
able to deduce properties of $\Gamma$ from properties of repeated
games of stage games $G'$ which are subgames of $G$. This can be
more naturally done when we consider a wider class of repeated
games which we call {\em stochastic repeated games}. Within this
class the repeated game of $G'$ can be imbedded in the repeated
game of $G$, thus enabling us to make the required deductions.

Let $S$ be a countable set of states which also includes an {\em
end state} $e$. We consider a game $\Gamma^S$ in which the game
$G$ is played repeatedly. Before each round a state from $S$ is
selected according to a probability distribution which depends on
the history of the previous terminal nodes and states. When the
state $e$ is realized the game ends. The selected state is known
to the players. The strategy played in each round depends on the
history of the terminal nodes and states. We now describe
$\Gamma^S$ formally.

\noindent {\bf Histories.} The set of   infinite histories in
$\Gamma^S$, is $H_\infty=(S\times Z)^\omega$. For $t\ge 0$ the set
of finite history of $t$ rounds,  is $H_t=(S\times Z)^t$, and the
set of preplay histories of $t$ rounds is $H_t^p=(S\times
Z)^t\times S$.  Denote $H=\cup_{t=0}^\infty H_t$ and
$H^p=\cup_{t=0}^\infty H_t\times S$. The subset of $H^p$ of
histories that terminate with $e$ is denoted by $F$. For $h\in
H_\infty$ and $t\ge 0$ we denote by $h_t$ the history in $H_t$
which consists of the first $t$ rounds in $h$. For finite and
infinite histories $h$ we denote by $\bar h$ the sequence of
terminal nodes in $h$.

\noindent {\bf Transition probabilities.} For each  $h\in H$,
$\tau(h)$ is a probability distribution on $S$. For $s\in S$,
$\tau(h)(s)$ is the probability of transition to state $s$ after
history $h$.  The probability that the game ends after $h$ is
$\tau(h)(e)$.

\noindent {\bf Super strategies.} After $t$ rounds the player
observes the history of $t$ pairs of a state and a terminal node,
and the state that follows them, and then plays $G$. Thus, a super
strategy for player $i$ is a function $\Sigma_i$ from
$H^p\setminus F$ to $i$'s strategies in $G$. We denote by
$\Sigma(h)(z)$ the probability of reaching terminal node $z$ when
$\Sigma(h)$ is played.

\noindent {\bf The super play distribution.} The super strategy
$\Sigma$ induces the {\em super play distribution} which is a
probability distribution $P$ over $H_\infty\cup F$. It is the
unique extension of the distribution over finite histories which
satisfies
\begin{equation}\label{playdistribution}
P(h,s)= P(h) \tau(h)(s)
\end{equation}
for $h\in H$, and
\begin{equation}\label{preplaydistribution}
P(h,z)= P(h)\Sigma(h)(z)
\end{equation}
for $h\in H^p$.

 \noindent {\bf The valuation super strategy.} Player
$i$'s valuation super strategy in $\Gamma^S$, starting with
valuation $v$, is the super strategy $\Sigma_i$ which satisfies
$\Sigma_i(h)=\sigma^{v^{\bar h}}$.

\subsection{Subgames}
We show now how a stochastic repeated game of a subgame of $G$ can
be imbedded in $\Gamma^S$.

For a  node $n$ in $G$, denote by $G_n$ the subgame starting at
$n$. Fix a super strategy profile $\Sigma$ in $\Gamma^S$ and the
induced super play distribution $P$ on $H_\infty$. In what follows
we describe a  stochastic super game $\Gamma_n^{S'}$, in which the
stage game is $G_n$. For this we need to define the state space
$S'$.  We tag histories and states in the game $\Gamma_n^{S'}$, as
well as terminal nodes in $G_n$. Our purpose in this construction
is to imbed $H'_\infty$ in $H_\infty$. The idea is to regard these
rounds in a history $h$ in $H_\infty$ in which node $n$ is not
reached as states in $S'$.

Let $S'$ be defined as the set of all  $h\in H^p$, such that node
$n$ is never reached in $h$. Obviously, $S'$ subsumes $S$, and in
particular includes the end state $e$. Note that the set
$H'_\infty$ of infinite history in $\Gamma_n^{S'}$ can be
naturally viewed as a subset of $H_\infty$, $H'$ as a subset of
$H$, and  ${H'}^p$ as a subset of $H^p$. We use this fact to
define the transition probability distribution $\tau'$ in
$\Gamma_n^{S'}$ as follows.

For any $s'\ne e$ in $S'$ and $h'\in H$  with $ P(h')>0$,
\begin{equation}\label{transitionsub}
  \tau'(h')(s')= P(h',s'\mid h')\Sigma(h',s')(n),
\end{equation}
where $\Sigma(h',s')(n)$ is the probability that node $n$ is
reached under the strategy profile $\Sigma(h',s')$. For $e$,
$\tau'(h')(e)= P(E\mid h')$, where $E$ consists of all histories
$h\in H_\infty\cup F$ with initial segment $h'$ such that $n$ is
never reached after this initial segment.

Note that $\tau'(h')(s')$ is the probability of all histories in
$H_\infty\cup F$ that start with $(h',s')$ and followed by a
terminal node of the game $G_n$. These events and the event $E$
described above, form a partition of $H_\infty\cup F$, and
therefore $\tau'$ is a probability distribution.

\begin{claim}\label{thesamedistribution}  Define a super strategy profile
$\Sigma'$ in $\Gamma_n^{S'}$, by
\begin{equation}\label{strategysub}
  \Sigma'(h')=\Sigma_n(h')
\end{equation}
for each $h'\in {H'}^p$, where the right-hand side  is the
restriction of $\Sigma(h')$ to $G_n$. Then, the restriction of $
P$ to $H'_\infty$ coincides with the super play probability
distribution $ P'$, induced by $\Sigma'$.
\end{claim}

\proof{} It is enough to show that $ P$ and $ P'$ coincide on
$H'$. The proof is by induction on the length of $h'\in H'$.
Suppose $P'(h')= P(h')>0$ and consider the history $(h,s',z')$.
Then, by the definition of the super play distribution
(\ref{playdistribution}) and (\ref{preplaydistribution}), $$
P'(h',s',z')= P'(h')\tau'(h')(s')\Sigma'(h',s')(z').$$ By the
induction hypothesis and the definitions of $\tau'$ in
(\ref{transitionsub}), the righthand side is
$P(h',s')\Sigma(h',s')(n)\Sigma'(h',s')(z')$. By the definition of
$\Sigma'$  in (\ref{strategysub}), this is just
$P(h',s')\Sigma(h',s')(n)\Sigma_n(h',s')(z').$ The right-hand
side, in turn, is just $P(h',s')\Sigma(h',s')(z')= P(h',s',z')$.
\qed

Next, we note that playing by valuation is inherited by subgames.

\begin{claim}\label{thesamestrategy}  Suppose that $i$'s strategy in
$\Gamma^S$, $\Sigma_i$, is  the valuation super strategy starting
with $v$, and using either the myopic strategy and the memoryless
revision rules, or the $\delta$-exploratory myopic strategy and
the averaging revision rules. Then the induced strategy in
$\Gamma_n^{S'}$, $\Sigma'_i$, is the valuation super strategy
starting with $v_{n}$
---the restriction of $v$ to the subgame $G_n$---and following the
corresponding rules.
\end{claim}
\proof{} The valuation super strategy in $\Gamma_n^{S'}$, starting
with $v_{n}$, requires that after history $h'\in H'$, strategy
$\sigma^{v_{n}^{\bar{h'}}}$ is played. Here, $\bar{h'}$ is the
sequence of all terminal nodes in $h'$, which consists of terminal
nodes in $G_n$. These are also all the terminal nodes of $G_n$, in
$h'$,  when the latter is viewed as a history in $H$.

When $h'$ is considered as a history in $H$, then the strategy
$\Sigma_i(h')$ is $\sigma^{v^{\bar h'}}$, where $\bar h'$ is the
sequence of all terminal nodes in  $h'$. $\Sigma'_i(h')$ is the
restriction of $\sigma^{v^{\bar h'}}$ to $G_n$. But along the
history $h'$, the valuation of nodes in the game $G_n$ does not
change in rounds in which terminal nodes which are {\em not} in
$G_n$ are reached. Therefore, $\Sigma'_i(h')$ and $\sigma^{v^{\bar
h'}}$ are the same. \qed

\subsection{Win-lose games}
The game $\Gamma$ is in particular a  stochastic repeated game,
where there is only one state, besides $e$, and transition to $e$
(that is, termination of the game) has null probability. We prove
all three theorems for the wider class of stochastic repeated
games. The theorems can be stated verbatim for this wider class of
games, with one obvious change: any claim about almost all
histories should be replaced by a corresponding claim for almost
all \emph{infinite} histories.

All the theorems are proved by induction on the depth of the game
$G$. The proofs for games of depth 0 (that is, games in which
payoffs are determined in the root, with no moves) are
straightforward and are omitted. In all the proofs,
$R=\{n_1,\dots, n_k\}$ is the set of all the immediate successors
of the root $r$.

\proof{ of Theorem \ref{main}} Assume that the claim of the
theorem  holds for all the subgames of $G$. We examine first the
case that the first player is not $i$. By the stipulation of the
theorem, player $i$ can guarantee payoff 1 in each of the games
$G_{n_j}$ for $j=1,\dots,k$.

Consider now the game $\Gamma_{n_j}^{S'}$, the super strategy
profile $\Sigma'$, and the induced super play distribution $ P'$.
By the induction hypothesis, and claim 2, for each $j$,  for $
P'$-almost all infinite histories  there is a time after which
player $i$ is paid 1.  In view of Claim 1, for $ P$-almost all
histories in $\Gamma^S$ in which $n_j$ is reached infinitely many
times, there exist a time after which player $i$ is paid 1,
whenever $n_j$ is reached. Consider now a nonempty subset $Q$ of
$R$. Let $E_Q$ be the set of infinite histories in $\Gamma^S$ in
which node $n_j$ is reached infinitely many times iff $n_j\in Q$.
Then, for $ P$-almost all histories in $E_Q$  there is a time
after which player $i$ is paid 1. The events $E_Q$ when $Q$ ranges
over all nonempty subsets of $R$, form a partition of the set of
all infinite histories, which completes the proof in this case.

Consider now the case that $i$ is the first player in the game. In
this case there is at least one subgame $G_{n_j}$ in which $i$ can
guarantee the payoff 1. Assume without loss of generality that
this holds for $j=1$.

For a history $h$ denote by $R^+_t$ the random variable that takes
as values the subset of the nodes in $R$ that have a positive
valuation after $t$ rounds. When $R^+_t$ is not empty, then $i$
chooses at $r$, with probability 1, one the nodes in $R^+_t$. As a
result the valuation of this node after the next round is 0 or 1,
while the valuation of  all other nodes does not change. Therefore
we conclude that $R^+_t$ is weakly decreasing when
$R^+_t\ne\emptyset$. That is, $ P(R^+_{t+1}\subseteq R^+_t\mid
R^+_t\ne\emptyset)=1$.

Let $E^+$ be the event that $R^+_t=\emptyset$ for only finitely
many $t$'s. Then, for $ P$-almost all histories in $E^+$ there
exists time $T$ such that $R^+_t$ is decreasing for $t\ge T$.
Hence,  for $ P$-almost all histories in $E^+$  there is a
nonempty subset $R'$ of $R$, and time $T$, such that $R^+_t=R'$
for $t\ge T$. But in order for the set of nodes in $R$ with
positive valuation not to change after $T$, player $i$ must be
paid 1 in each round after $T$. Thus we only need to show that $
P(\bar{E^+})=0$.

Consider the event $E^1$ that $n_1$ is reached in infinitely many
rounds. As proved before by the induction hypothesis, for $
P$-almost all histories in $E^1$, there exists $T$, such that the
valuation of $n_1$ is 1, for each round  $t\ge T$ in which $n_1$
is reached. The valuation of this node does not change in rounds
in which it is not reached. Thus, $E^1\subseteq E^+$ $ P$-almost
surely.

We conclude that for  $ P$-almost all histories in $\bar{E^+}$
there is a time $T$, such that $n_1$ is not reached after time
$T$. But  $ P$-almost surely for such histories there are
infinitely many $t$'s in which the valuation of all nodes in $R$
is 0. In each such history, the probability that $n_1$ is not
reached is $1-1/k$, which establishes $ P(\bar{E^+})=0$. \qed

\proof{ of Theorem \ref{generic}} Let $i$ be the player at the
root of  $G$. By the induction hypothesis and Claim 1, for each of
the supergames $\Gamma_{n_j}^{S'}$, $j=1,\dots,k$, for $P'$-almost
infinite histories in this super game, there is a time after which
the same terminal node is reached. By Claim 2, for $P$-almost all
histories of $\Gamma$ in which $n_j$ recurs infinitely many times
there is a time after which $i$'s valuation of this node is
constantly the payoff of  the same terminal node of $G_{n_j}$.

It is enough that we show that for $P$-almost all infinite
histories in $\Gamma^S$, there is a time after which the same node
from $R$ is selected with probability 1 at the root. Suppose that
this is  not the case. Then there must be a set of histories $E$
with $P(E)>0$, two nodes $n_j$ and $n_l$, and two terminal nodes
$z_j$ and $z_l$ in $G_{n_j}$ and $G_{n_l}$ correspondingly, that
recur infinitely many times in this set. Therefore, for $P$-almost
all histories in $E$, $i$'s valuation of $n_j$ and $n_l$ is
$f_i(z_j)$ and $f_i(z_l)$. Since $G$ is generic, we may assume
that $f_i(z_j)>f_i(z_l)$. Thus, for $P$-almost all histories in
$E$, there is a time after which the conditional probability of
$n_l$ given the history is 0. Which is a contradiction. \qed

\subsection{The case of payoff functions with more than two
values} We prove Theorem \ref{maxmin} for stochastic repeated
games, where the conclusion of the theorem holds for
$P_\delta$-almost all {\em infinite} histories.

\proof{ of Theorem \ref{maxmin}}    Assume that the claim holds
for all the subgames of $G$.  We denote by $\rho_j$, $i$'s
individually rational (maxmin) payoff in $G_{n_j}$.

We denote by $\bar f^t(h)$, $i$'s average payoff at time $t$ in
history $h$. Fix a subgame $G_{n_j}$. Histories in the game
$\Gamma^{S'}_{n_j}$ are  tagged. Thus, $\bar f^t(h')$ is $i$'s
average payoff at time $t$ in history $h'$ in $\Gamma^{S'}_{n_j}$.

Let $h$ be a history in $\Gamma$ in which $n_j$ recurs infinitely
many times at $t_1,t_2,\dots$. Let $\bar h=(z_1,z_2,\dots)$.
Denote by $\bar f_j^{t}(h)$ $i$'s average payoff until $t$
\emph{at the times $n_j$ was reached}, that is, $$\bar f_j^{t}(h)=
\frac{1}{|\{l: t_l<t\}|}\sum_{l: t_l<t} f(z_{t_l}).$$

The history $h$ can be viewed as an infinite history $h'$ in
$\Gamma^{S'}_{n_j}$. Moreover, for each $l$, $\bar f^{l}(h')=\bar
f_j^{t_l}(h)$. By the definition of $\bar f_j^{t}(h)$, it follows
that if there exists $L$ such that for each $l>L$,  $\bar f^l(h')>
\rho_j-\e$, then there exits $T$ such that for each $t>T$, $\bar
f_j^{t}(h)> \rho_j-\e$. By the induction hypothesis there is
$\d_0$, such that for all $0<\d<\d_0$, for $P'_\d$-almost all
histories $h'$ there exists such an $L$. Thus, by Claims
\ref{thesamedistribution} and \ref{thesamestrategy}, there exists
$\d_0$, such that for all $j$ and $0<\d<\d_0$, for $P_\d$-almost
all histories $h$ in $\Gamma^S$ in which $n_j$ recurs infinitely
many times, there exists a time $T$ such that for each $t>T$,
$\bar f_j^t(h)> \rho_j-\e$.

We examine first the case that the first player is not $i$.
Obviously, in this case, $\rho=\min_j\rho_j$.

Let $Q$ be a nonempty  subset of $R$, and let $E_Q$ be the set of
all infinite histories in which the set of nodes that recurs
infinitely many times is $Q$. Consider a history $h$ in $E_Q$,
with $\bar h=(z_1,z_2,\dots)$. Let $\n_j^t(h)$ be the number of
times $n_j$ is reached in $h$ until time $t$.  Then,  $$\bar
f^{t}(h) = \frac{1}{t}\sum_{j=1}^k \n_j^t(h) \bar f_j^{t}(h)\ge
\min_{j:\, n_j\in Q} \bar f_j^{t}(h),$$ where the inequality
holds, because $\sum_j \n_j^t(h) =t$, and for $j\notin Q$,
$\n_j^t(h)= 0$.
 Thus for $P_\delta$-almost all histories $h$ in $E_Q$,
\begin{equation*}
\begin{split}
\varliminf_{t\rightarrow\infty}\, \bar
f^{t}(h)&\ge\varliminf_{t\rightarrow\infty} \min_{j:\, n_j\in Q}
\bar f_j^{t}(h)\\ &\ge \min_{j:\, n_j\in
Q}\varliminf_{t\rightarrow\infty}\,\bar f_j^{t}(h)\\&
> \min_{j:\, n_j\in Q}\rho_j-\e\\ & \ge\rho-\e.
\end{split}
\end{equation*}
Since this is true for all $Q$, the conclusion of the theorem
follows for all infinite histories.

Next, we examine  the case that $i$ is the first player. Note that
in this case, for each node $n_j$, $\bar f_j^{t}(h)=v^{h_t}(n_j)$.
Observe, also, that  for $P_\d$-almost all infinite histories $h$
in $\Gamma^S$, each of the subgames $G_{n_j}$ recurs infinitely
many times in $h$. Indeed, after each finite history, each of the
games $G_{n_j}$ is selected by $i$ with probability $\delta$ at
least. Thus, the event that one of these games is played only
finitely many times has probability 0.

Let $X_t$ be a binary random variable over histories such that
$X_t(h)=1$ for histories $h$ in which the node $n_{j_0}$ selected
by player $i$ at time $t$ satisfies,
\begin{equation}\label{jzero}
v^{h_t}(n_{j_0})> \rho-\e/2,
\end{equation}
and $X_t=0$ otherwise.

\begin{claim}\label{epsilon1}
There exists $\d_0$ such that for all $j=1\dots k$ and any
$0<\d<\d_0$,  for $P_\d$-almost all infinite histories $h$ in
$\Gamma^S$ there is time $T$ such that for all $t>T$,
\begin{equation}\label{convergence}
v^{h_t}(n_j)> \rho_j-\e/4,
\end{equation}
\begin{equation}\label{nextround}
|v^{h_t}(n_j)-v^{h'_{t+1}}(n_j)|<\e/4,
\end{equation}
for each history $h'$ such that $h'_t=h_t$, and
\begin{equation}\label{expectation}
E_\d(X_{t+1}|h_t)\ge 1-\d,
\end{equation}
where $E_\d$ is the expectation with respect to $P_\d$.
\end{claim}

The inequality (\ref{convergence}) follows from the induction
hypothesis. For (\ref{nextround}), note that if $n_j$ is not
reached in round $t+1$ then the difference in (\ref{nextround}) is
0. If $n_j$ is reached then $v^{h'_{t+1}}=\big(\n v^{h_t}(n_j) +
f(z_{t+1})\big)/(\n +1)$, where $\n$ is the number of times $n_j$
was reached in $h_t$ and $f(z_{t+1})$ is the payoff in round
$t+1$. But, $\n$ goes to infinity with $t$, and thus
(\ref{nextround}) holds for large enough $t$.

For (\ref{expectation}), observe that (\ref{convergence}) implies
$\max_j v^{h_t}(n_j)> \rho-\e/4$, as $\rho=max_j \rho_j$.  Then,
by (\ref{nextround}), $\max_j v^{h'_{t+1}}(n_j)> \rho-\e/2$ for
each history $h'$ such that $h'_t=h_t$. Therefore, after $h_t$,
player $i$ chooses, with probability at least $\d$, a node
$n_{j_0}$ that satisfies (\ref{jzero}), which shows
(\ref{expectation}).

The information about the conditional expectations in
(\ref{expectation}) has a simple implication for the averages of
$X_t$. To see it we use the following convergence theorem from
Lo\`eve (1963) p. 387.

\begin{stability}
Let $X_t$ be a sequence of random variables with variance
$\sigma^2_t$. If
\begin{equation}\label{variances}
\sum_{t=1}^\infty \sigma^2_t/t^2< \infty,
\end{equation}
 then
\begin{equation}\label{stability}
\bar X_t- \frac{1}{t}\sum_{l=1}^t E(X_l\mid
 X_1,\dots,X_{l-1}))\rightarrow 0,
\end{equation}
almost surely, where $\bar X_t=(1/t)\sum_{l=1}^t X_l$.
\end{stability}

Consider now the restriction of the random variables $X_t$ to the
set of infinite histories with $P_\d$ conditioned on this space.
From (\ref{expectation}) it follows that on this space, almost
surely $\varliminf_{t\rightarrow\infty}\,\frac{1}{t}\sum_{l=1}^k
E(X_l\mid h_l))\ge 1-\d$. Therefore, almost surely
$\varliminf_{t\rightarrow\infty}\,\frac{1}{t}\sum_{l=1}^k
E(X_l\mid X_1,\dots,X_{l-1}))\ge 1-\d$.  This is so, because the
field generated by the the random variables $(X_1,\dots, X_{l-1})$
is coarser than the field generated by histories $h_t$.  Since
condition (\ref{variances}) holds for $X_t$, it follows by the
Stability Theorem that for $P_\d$-almost all infinite histories
$h$,
\begin{equation}\label{limit}
\varliminf_{t\rightarrow\infty}\, \bar X_t\ge 1-\d.
\end{equation}

By the definition of $X_t$, $$\bar f^{t}(h) =
\frac{1}{t}\sum_{j=1}^k \n_j^t(h) v^{h_t}(n_j)\ge \bar
X_t(h)(\rho-\e/2) + (1-\bar X_t(h))\underbar M,$$ where $\underbar
M$ is the minimal payoff in $G$. If we choose $\d_0$ such that
$(1-\d_0)(\rho-\e/2) +\d_0\underbar M > \rho-\e$, then by
(\ref{limit}), for each $\d<\d_0$,
$\varliminf_{t\rightarrow\infty}\, f^{t}(h)> \rho-\e$ for
$P_\d$-almost all infinite histories. \qed

The proof of Theorem \ref{perfect} is also extended to stochastic
repeated games. We show that the conclusion of the theorem holds
for $P_\d$-almost all infinite histories.

\proof{ of Theorem \ref{perfect}} Assume that the claim of the
theorem holds for all the subgames of $G$.  We denote by $v_j$ the
restriction of the valuation $v$ to $G_{n_j}$, and by $\b_{i,j}$,
$i$'s perfect equilibrium strategy there, which is also the
restriction of $\b_i$ to this game.

\begin{claim}
Let $i_0$ be the player at the root,  $\pi_j$ be $i_0$'s payoff in
the perfect equilibrium of $G_{n_j}$, and $\e>0$.

Then there exists $\d_0>0$ such that for all $0<\d<\d_0$, node
$n_j$, and player $i$, for $P'_\d$ almost all infinite histories
$h'$ of $\Gamma^{S'}_{n_j}$ there exists $T$ such that for all
$t>T$,
\begin{equation}\label{perfect-induction}
  \s_i^{v^{h'_t}_{j}}(m)=(1-\d)\beta_{i,j}(m)+\d\m(m)
\end{equation}
for each node $m\in M_i$ in $G_{n_j}$, and
\begin{equation}\label{perfect-expectation}
|E_\d(f_j^{t+1}|h'_t) -\pi_j|<\e
\end{equation}
where $E_\d$ is the expectation with respect to $P'_\d$, and
$f_j^{t+1}$ is $i$'s payoff in round $t+1$.
\end{claim}

The equality (\ref{perfect-induction}) is the induction
hypothesis. Consider a history $h'_t$ for  which
(\ref{perfect-induction}) holds. In the round that follows $h'_t$,
the  perfect  equilibrium path in $G_{n_j}$ is played with
probability $(1-\d)^{d-1}$ at least, where $d$ is the depth of
$G$. Player $i_0$'s payoff in this path is $\pi_j$. Thus for small
enough $\d_0$, (\ref{perfect-expectation}) holds.

By Claims \ref{thesamedistribution} and \ref{thesamestrategy} it
follows from (\ref{perfect-induction}) that for $0<\d<\d_0$,  for
$P_\d$ all histories $h$ in $\Gamma$, there exists $T$ such that
for all $t>T$  the strategies played in each of  the games
$\Gamma^{S'}_{n_j}$ is the perfect equilibrium of $G_{n_j}$. Thus,
to complete the proof it is enough to show that in addition, at
the root, $i_0$ chooses in these rounds, with probability $1-\d$,
the node $n_{j_0}$ for which $\b_{i_0}(r)=n_{j_0}$. For this we
need to show that $i_0$'s valuation of $n_{j_0}$ is higher than
the valuation of all other nodes $n_j$.

To show it,  let $3\e$ be the difference between $\pi_{j_0}$ and
the second highest payoffs $\pi_j$. By the assumption of the
uniqueness of the perfect equilibrium, $\e>0$. Note that as all
players' strategies are fixed for $t>T$,
$\lim_{t\rightarrow\infty} \frac{1}{t}\sum_{l=1}^t
E_\d(f_{i_0}^{t+1}|h'_t)$ exists. Using the stability Theorem, as
in Theorem \ref{maxmin}, we conclude that
$\lim_{t\rightarrow\infty} \bar f_j^{t}(h')$ exists, and by
(\ref{perfect-expectation}) the inequality
$|\lim_{t\rightarrow\infty} \bar f_j^{t}(h') -\pi_j|<\e$ holds,
where $\bar f_j^{t}(h')$ is $i_0$'s average payoff until round $t$
of history $h'$, in the game $\Gamma^{S'}_{n_j}$.

As in the proof of Theorem \ref{maxmin}, it follows that for
$P_\d$-almost all infinite histories $h$ in $\Gamma$,
$|\lim_{t\rightarrow\infty} v^{h_t}(n_j) -\pi_j|<\e$. But then,
for $P_\d$-almost all infinite histories $h$ there exists $T$ such
that for all $t>T$, $v^{h_t}(n_{j_0})$ is the highest valuation of
all the nodes $n_j$. \qed

\end{document}